\documentclass[10pt]{article} 
\usepackage[preprint]{tmlr}
\renewcommand{\headrulewidth}{0pt}

\usepackage{amsmath,amsfonts,bm}

\def\eqref#1{equation~\ref{#1}}

\def\1{\bm{1}}

\DeclareMathAlphabet{\mathsfit}{\encodingdefault}{\sfdefault}{m}{sl}
\SetMathAlphabet{\mathsfit}{bold}{\encodingdefault}{\sfdefault}{bx}{n}

\usepackage{hyperref}
\hypersetup{hidelinks,
  pdftitle={OpenMidnight: Closing the Performance Gap Between Open and Proprietary Pathology Foundation Models},
  pdfauthor={Ratna Sagari Grandhi, Anis Chihoub, Ryan Kim, Daniel Z. Kaplan, Chao Feng, Anish Dulal, Leema Krishna Murali, Rohitkumar Datchanamourty, Nevasini Sasikumar, Debojyoti Das, Ahmed Hassayoune, Rhizo Nymph, Anas Zafar, Connor Lane, Benjamin Warner, Hassan Muhammad, Tanishq Mathew Abraham, Paul S. Scotti}
}
\usepackage{xurl}
\usepackage{booktabs}
\usepackage{graphicx}
\usepackage{float}
\usepackage{placeins}
\usepackage{pdflscape}
\usepackage{xcolor}
\usepackage{colortbl}
\usepackage{enumitem}

\title{OpenMidnight: Closing the Performance Gap Between Open and Proprietary Pathology Foundation Models}

\newcommand{\preprintauthors}{\normalfont
\begin{minipage}{\textwidth}
\centering\small
\textbf{Ratna Sagari Grandhi}\textsuperscript{1}\quad
\textbf{Anis Chihoub}\textsuperscript{2,1}\quad
\textbf{Ryan Kim}\textsuperscript{3,1}\quad
\textbf{Daniel Z. Kaplan}\textsuperscript{4,1}\\[0.3em]
\textbf{Chao Feng}\textsuperscript{1}\quad
\textbf{Anish Dulal}\textsuperscript{5,1}\quad
\textbf{Leema Krishna Murali}\textsuperscript{6,1}\quad
\textbf{Rohitkumar Datchanamourty}\textsuperscript{1}\\[0.3em]
\textbf{Nevasini Sasikumar}\textsuperscript{7,1}\quad
\textbf{Debojyoti Das}\textsuperscript{1}\quad
\textbf{Ahmed Hassayoune}\textsuperscript{1}\quad
\textbf{Rhizo Nymph}\textsuperscript{1}\\[0.3em]
\textbf{Anas Zafar}\textsuperscript{8,1}\quad
\textbf{Connor Lane}\textsuperscript{9,1}\quad
\textbf{Benjamin Warner}\textsuperscript{9,1}\\[0.3em]
\textbf{Hassan Muhammad}\textsuperscript{9,1}\quad
\textbf{Tanishq Mathew Abraham}\textsuperscript{9,1}\quad
\textbf{Paul S. Scotti}\textsuperscript{9,1}\\[0.3em]
\par\smallskip
{\footnotesize
\textsuperscript{1}\href{https://medarc.ai}{\underline{MedARC}}\quad
\textsuperscript{2}Rutgers University\quad
\textsuperscript{3}University of Illinois at Urbana-Champaign\\
\textsuperscript{4}realiz.ai\quad
\textsuperscript{5}University of Oregon\quad
\textsuperscript{6}UCLA\quad
\textsuperscript{7}UC San Diego\\
\textsuperscript{8}University of Texas, MD Anderson Cancer Center\quad
\textsuperscript{9}\href{https://sophont.med}{\underline{Sophont}}
\par}
\smallskip
{\small Correspondence: \href{mailto:paul@sophont.med}{\nolinkurl{paul@sophont.med}}\par}
\end{minipage}}
\author{\preprintauthors}

\begin{document}
\raggedbottom
\maketitle

\begin{abstract}
Pathology foundation models offer transferable representations for a range of downstream tasks in computational pathology, but reliance on proprietary data, substantial compute requirements, and limited availability of training code and preprocessing pipelines hinder independent validation and further methodological development. We present OpenMidnight 1.1, an open-source replication and refinement of Midnight trained on approximately 12,000 publicly available whole-slide images from The Cancer Genome Atlas. Using an encoder initialized with pretrained DINOv2 weights, OpenMidnight 1.1 uses approximately 320 H100 GPU-hours for pathology pretraining---a 34-fold reduction relative to the compute reported for Midnight. Across 16 classification and four segmentation datasets in THUNDER, OpenMidnight 1.1 achieves the highest mean segmentation Dice (69.7) and ranks fourth among 14 models, ahead of Midnight, with performance that matches or exceeds models trained on substantially larger datasets in several tasks. Our ablation studies additionally examine how initialization, data volume, model size, and tissue composition affect downstream performance, providing empirical evidence to inform the development of future training recipes. Our release of the training code, preprocessing pipeline, and model weights lowers barriers to reproducible research and further development of competitive pathology foundation models.
\end{abstract}

\section{Introduction}
In recent years, self-supervised learning has transformed computational pathology by enabling foundation models to learn from terabytes of histopathology images without costly, time-consuming expert annotation, a major bottleneck for modeling gigapixel whole-slide images (WSIs) \citep{ochi2025pathology}. However, generalizability, interpretability, and reproducibility remain important concerns for their integration into clinical workflows in the real world \citep{fda2026generativeai}.

Open access to training code, preprocessing pipelines, pretraining data, and model weights supports independent evaluation and reproducible development~\citep{heil2021reproducibility,wagner2024built}. Yet leading models, including GigaPath, Virchow2, H-Optimus-0, and Hibou, have largely relied on proprietary datasets and substantial computational resources, often without releasing complete pretraining code~\citep{xu2024gigapath,zimmermann2024virchow2,saillard2024hoptimus0,nechaev2024hibou}. These barriers restrict the research community's ability to reproduce and extend their results. Meanwhile, independent benchmarks suggest that larger pretraining datasets do not consistently translate into better downstream performance~\citep{marza2025thunder,gatopoulos2024eva,ma2025pathbench}. Together, these observations motivate more efficient and reproducible training approaches built on publicly available data.

Midnight~\citep{karasikov2025midnight} demonstrated the potential of this philosophy, achieving performance comparable to models trained on more than a million slides using only 12,000 publicly available whole slide images (WSIs) from The Cancer Genome Atlas (TCGA)~\citep{weinstein2013tcga}. Nevertheless, its training required 32 H100 GPUs for 14 days. Its weights were released under a non-commercial license, but its training code was not made available, limiting independent replication. Contributions of pretraining initialization, model size, and pretraining data scale and composition also remained insufficiently characterized, leaving open whether a more accessible training recipe could facilitate further improvement of performance.

\begin{figure}[!htbp]
    \centering
    \includegraphics[width=0.8\textwidth]{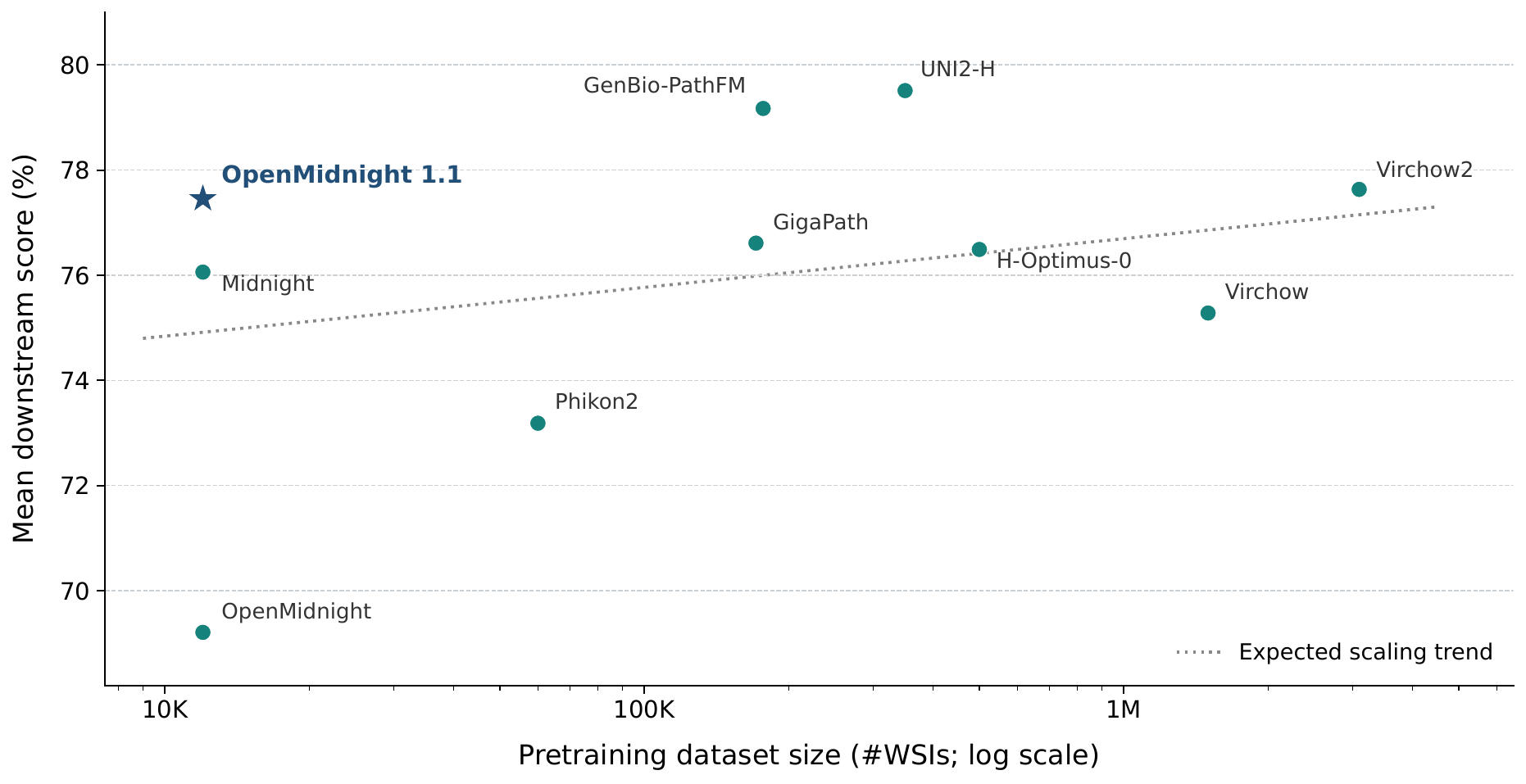}
    \caption{\textbf{Benchmark performance relative to pretraining data scale.} Scores equally average the four THUNDER category means: linear probing, KNN, and 16-shot classification (F1), and segmentation (Dice). Each classification category averages 16 datasets, and segmentation averages four. Pretraining slide counts are shown on a logarithmic scale. OpenMidnight 1.1, trained on approximately 12,000 publicly available TCGA slides, performs comparably to models trained on substantially larger slide collections.}
    \label{fig:perf_overview}
\end{figure}

Here, we present OpenMidnight 1.1, a fully open-source replication and refinement of Midnight trained on approximately 12,000 publicly available TCGA slides. OpenMidnight 1.1 outperforms Midnight on the THUNDER benchmark suite~\citep{marza2025thunder} using a fraction of Midnight's reported training compute, at an estimated training cost of \$800. It ranks fourth among 14 models across five THUNDER evaluation categories, remaining competitive with models trained on one to two orders of magnitude more data (Figure~\ref{fig:perf_overview}). Controlled ablations assess the contributions of DINOv2 initialization~\citep{oquab2023dinov2}, model size, pretraining data volume and corpus composition; additional analyses examine the geometry of the learned representations and qualitatively assess prediction errors in a downstream segmentation task. We release the complete training pipeline, preprocessing scripts, and model weights to support reproducibility and lower the barrier to developing pathology foundation models aimed at clinical deployment.

\section{Related Work}

Pathology foundation models learn transferable representations from histopathology images, typically through self-supervised pretraining on tiles extracted from whole-slide images. Early models demonstrated the value of domain-specific pretraining using publicly accessible data: CTransPath employed contrastive learning~\citep{wang2022ctranspath}, while Phikon adopted the iBOT masked-image modeling objective~\citep{filiot2023phikon}. More recent models, including GigaPath, Virchow2, H-Optimus-0, and Hibou, have adopted DINOv2-based training~\citep{oquab2023dinov2} and substantially expanded their pretraining corpora, often incorporating proprietary data~\citep{xu2024gigapath,zimmermann2024virchow2,saillard2024hoptimus0,nechaev2024hibou}. Although these models achieve strong downstream performance, independent reproduction remains difficult when their pretraining data or training code are not publicly available.

Midnight demonstrated that competitive pathology foundation models could be trained using approximately 12,000 publicly available TCGA slides~\citep{karasikov2025midnight}. Although its model weights were released, the training code was not publicly available, and pretraining required substantial computational resources. Midnight is the closest methodological predecessor to our work. We extend this direction by providing an open training pipeline, reducing pretraining compute, and examining the effects of initialization, data volume, model size, and tissue composition under the tested training configurations.

Independent benchmarks provide complementary assessments of pathology foundation models. EVA supports tile- and slide-level classification and nuclei segmentation~\citep{gatopoulos2024eva}, while HEST evaluates gene-expression prediction from histological morphology~\citep{jaume2024hest}. THUNDER broadens tile-level evaluation to include classification, segmentation, image retrieval, representation comparisons, and assessments of uncertainty and robustness~\citep{marza2025thunder}. Together, these frameworks support evaluation across different uses and properties of learned representations. Our study uses THUNDER to compare the released model with existing foundation models and EVA and HEST to evaluate training ablations.

\section{Methods}

OpenMidnight~\citep{kaplan2025openmidnight} was first introduced as an open-source replication and refinement of Midnight~\citep{karasikov2025midnight}, including both training code and the model weights, with closely matching performance. We retained key elements of training methods described in Midnight, including TCGA data, a DINOv2-initialized ViT-G/14 backbone, the self-supervised objective, tissue filtering, stain-aware augmentation and use of a kernel density estimation (KDE) regularizer~\citep{wang2020kde} in place of DINOv2's KoLeo regularizer. We changed how training patches were supplied: rather than sampling new regions from whole-slide images throughout training, we selected random, tissue-filtered patch locations in advance. We also moderated color augmentation and trained for 250,000 steps on eight H100 GPUs, compared with Midnight’s one million steps on 32 H100 GPUs. 

\begin{figure}[!htbp]
    \centering
    \includegraphics[width=\linewidth,
        height=0.4\textheight,
        keepaspectratio]{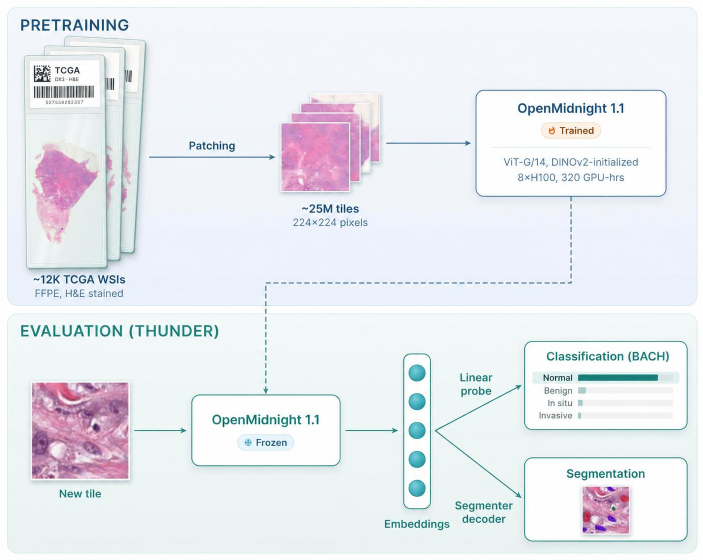}
    \caption{\textbf{Overview of the OpenMidnight 1.1 pipeline.} \textbf{Pretraining:} $\sim$25M tiles are extracted from $\sim$12K TCGA whole-slide images and used to train a DINOv2-initialized ViT-G/14 on 8 H100 GPUs (320 GPU-hours). \textbf{Evaluation:} the frozen model produces embeddings for THUNDER classification tasks and segmentation tasks.}
    \label{fig:architecture}
\end{figure}

OpenMidnight 1.1 introduces several updates and refinements to our initial blog release, including a correction to the sign of the KDE regularizer, a shorter training schedule, antialiased positional-embedding interpolation, and a checkpoint averaging procedure described below. The shortened 137,500-step schedule further reduces pretraining compute to 320 H100 GPU-hours—approximately 34-fold fewer than reported for Midnight. Figure~\ref{fig:architecture} summarizes the OpenMidnight 1.1 pretraining and evaluation pipeline.

\subsection{Model Architecture and Training Objective}

OpenMidnight 1.1 uses a ViT-G/14 encoder with approximately 1.1 billion parameters and four register tokens. We initialize the student encoder from Meta’s pretrained DINOv2 weights~\citep{oquab2023dinov2}, resizing the learned patch positional embeddings from a 37 × 37 grid to the 16 × 16 grid used for our 224 × 224-pixel global crops. Following DINOv2, separate DINO and iBOT projection heads each produce scores over 131,072 prototypes. Their MLP weights are initialized from a truncated normal distribution with standard deviation 0.02, and their biases are set to zero; the teacher is then initialized as a copy of the student.

During pretraining, the DINO loss matches the student’s class-token predictions from global and local crops to the teacher’s predictions from global crops of the same image. The iBOT loss asks the student to match the teacher’s prototype scores at patch locations hidden from the student but visible to the teacher. The teacher is updated as an exponential moving average of the student. Following Midnight, we also replace DINOv2’s KoLeo regularizer with a KDE regularizer~\citep{wang2020kde}, weighted by 0.05. The initial OpenMidnight implementation contained a sign error in this term: when added to the minimized training loss, it favored the opposite of the intended feature-spreading behavior. OpenMidnight 1.1 corrects the sign.

\subsection{Pretraining Data and Patch Sampling}

\begin{figure}[!htbp]
    \centering
    \includegraphics[width=0.85\textwidth]{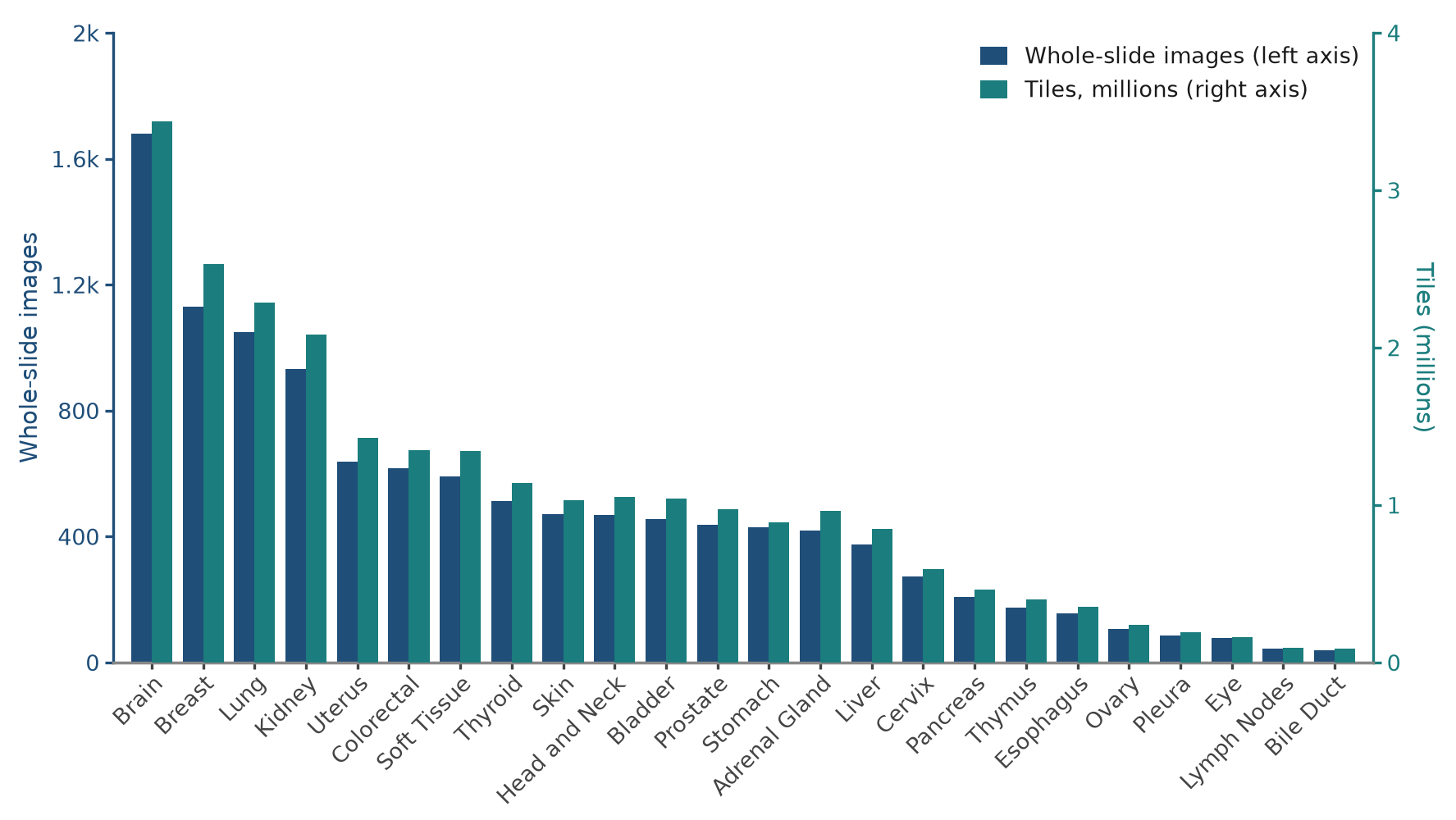}
     \caption{\textbf{Pretraining data distribution across primary sites.} Numbers of WSIs (navy, left axis) and sampled tiles (teal, right axis) in the pretraining corpus, comprising approximately 12k slides from 31 TCGA projects and 25 million sampled tiles, grouped by their primary sites.}
    \label{fig:data_distribution}
\end{figure}

Following Midnight, we use approximately 12,000 TCGA~\citep{weinstein2013tcga} WSIs spanning 31 cancer types and 24 primary sites. Before training, we select random locations for 224 × 224-pixel tiles across the slides’ available resolution levels, allowing sampled regions to overlap. We retain Midnight’s HSV-based tissue filter, accepting tiles for which more than 60\% of pixels fall within the specified tissue color range. This yields a pool of approximately 25 million sampled tiles. Figure~\ref{fig:data_distribution} shows the numbers of slides and tiles by primary site.

\subsection{Training Setup}
\label{sec:training_setup}

We train on eight 80 GB NVIDIA H100 GPUs with a batch size of 48 per GPU (384 in total), using AdamW for 137,500 optimizer steps. Following DINOv2’s square-root batch-size scaling, a base learning rate of \(3\times10^{-4}\) at a reference batch size of 1,024 gives a peak rate of approximately \(1.84\times10^{-4}\) for our batch size. The learning rate warms up to this value over 12,500 steps and then decays on a cosine schedule. Over the full run, weight decay increases from 0.04 to 0.2 and teacher momentum increases from 0.994 to 1.0, both on cosine schedules. We use a stochastic depth rate of 0.4.

Each tile produces two 224 × 224-pixel global crops and eight 98 × 98-pixel local crops. Following Midnight, we first apply HED stain augmentation to each crop with probability 0.5 and a perturbation range of ±0.05 per channel, before applying RGB color jitter. To avoid excessively strong color perturbations in histopathology tiles, we reduced the standard DINOv2 RGB color-jitter strengths for brightness and contrast from 0.4 to 0.2, saturation from 0.2 to 0.1, and hue from 0.1 to 0.05. We also omit solarization from the second global crop. For local crops, OpenMidnight 1.1 interpolates positional embeddings using bicubic interpolation with antialiasing and zero offset.

The run processes approximately 52.8 million tile presentations from the approximately 25-million-tile pool and takes about 40 hours, or 320 GPU-hours. We obtain the reported model by averaging the parameters of five teacher checkpoints saved at steps 87,500, 100,000, 112,500, 125,000, and 137,500.

\subsection{Evaluation Protocol}

We evaluate OpenMidnight 1.1 using THUNDER~\citep{marza2025thunder} across 16 classification and four segmentation datasets. For classification, we evaluate linear probing, KNN, and 16-shot classification using frozen image embeddings and report F1 scores. For semantic segmentation, we train THUNDER’s Segmenter decoder on frozen patch-token embeddings and report Dice scores. We also assess prediction calibration using expected calibration error (ECE). All evaluations follow THUNDER’s task-specific protocols, with computational modifications described below.

We evaluate the ablation runs using EVA~\citep{gatopoulos2024eva}, which covers tile- and slide-level classification and nuclei segmentation. With the encoder frozen, we follow EVA’s task-specific evaluation protocols and report balanced accuracy for classification and Dice score for segmentation. For gene-expression prediction, we use HEST~\citep{jaume2024hest} and report the mean Pearson correlation across its nine benchmark tasks.

THUNDER and HEST evaluations use an optimized pipeline adapted from their official codebases. The optimizations include FP16 encoder inference, feature caching, vectorized evaluation of frozen features, INT8 storage of patch-token features with FP16 scaling factors, and TF32 matrix multiplication in segmentation decoders. These optimizations reduce computation and storage requirements, but changes in numerical precision and implementation can affect the resulting scores. The optimized evaluation code is publicly available at \url{https://github.com/MedARC-AI/pathfm-full-evals}.

\section{Results}

\subsection{Benchmark Performance}

Table~\ref{tab:thunder_main} presents performance across five THUNDER evaluation categories, with classification and calibration evaluated on 16 datasets and segmentation evaluated on four datasets. Complete per-dataset results are provided in Appendix~\ref{app:thunder_full}. OpenMidnight 1.1 achieves the highest mean segmentation Dice (69.7) among all 14 models evaluated and ranks fourth overall with a rank sum of 25. It also achieves higher mean linear-probing F1 than models trained on substantially larger datasets, including H-Optimus-0 and Virchow2, which use more than 500,000 and approximately 3.1 million slides, respectively. These results are obtained using approximately 12,000 publicly available TCGA slides, reinforcing Midnight’s finding that competitive pathology foundation models can be trained on substantially smaller datasets.

OpenMidnight 1.1 exceeds Midnight in linear probing, KNN, few-shot classification, and segmentation, achieving a lower rank sum (25 versus 26) despite its higher calibration error (ECE of 5.8 versus 2.9). Both models use TCGA slides, but Midnight reports training on 32 H100 GPUs for 14 days, compared with eight H100 GPUs for approximately 40 hours for OpenMidnight 1.1—approximately 34-fold fewer reported GPU-hours. This comparison demonstrates competitive benchmark performance with substantially less pretraining compute. 

\begin{table}[!htbp]
\centering
\caption{\textbf{Performance on THUNDER.} Results are averaged over 16 classification datasets for linear probing, KNN, 16-shot classification, and calibration, and four segmentation datasets. Classification performance is measured by F1, calibration by ECE, and segmentation by Dice, all expressed as percentages. Models are ordered by the sum of their ranks across the five metrics. Bold indicates the best scores for each task type, and OpenMidnight 1.1 is highlighted. Adversarial attack results are excluded because numerical effects in our optimized evaluation pipeline can weaken attacks unevenly across models, compromising comparisons of adversarial robustness.}
\label{tab:thunder_main}
\small
\setlength{\tabcolsep}{5pt}
\renewcommand{\arraystretch}{1.05}
\resizebox{\linewidth}{!}{%
\begin{tabular}{lcc|ccccc|c}
\toprule
\textbf{Model} & \textbf{\#WSIs} & \textbf{Source} & \textbf{Lin. Prob.} $\uparrow$ & \textbf{Calib.} $\downarrow$ & \textbf{KNN} $\uparrow$ & \textbf{Few-shot} $\uparrow$ & \textbf{Seg} $\uparrow$ & \textbf{Rank sum} $\downarrow$ \\
\midrule
\rowcolor{gray!10}
\multicolumn{9}{l}{\textit{Histopathology models}} \\
\midrule
UNI2-H & 350K & Private & \textbf{85.7} & 3.9 & 83.3 & \textbf{79.8} & 69.2 & \textbf{11} \\
\rowcolor{gray!10}
GenBio-PathFM & 177K & HistAI, TCGA, GTEx, REG & 85.4 & 4.8 & \textbf{83.7} & 79.4 & 68.2 & 18 \\
Virchow2 & 3.1M & Private & 84.7 & 4.0 & 82.9 & 74.0 & 68.9 & 24 \\
\rowcolor{blue!15}
\textbf{OpenMidnight 1.1} & 12K & TCGA & 85.2 & 5.8 & 80.8 & 74.0 & \textbf{69.7} & 25 \\
Midnight & 12K & TCGA & 84.7 & \textbf{2.9} & 80.0 & 71.5 & 68.0 & 26 \\
\rowcolor{gray!10}
H-Optimus-0 & $>500$K & Private & 83.8 & 4.0 & 81.4 & 76.2 & 64.6 & 28 \\
H0-mini & 6.1K & TCGA\textsuperscript{\dag} & 83.7 & 3.8 & 79.7 & 75.0 & 68.2 & 29 \\
\rowcolor{gray!10}
GigaPath & 171K & Private & 83.6 & 3.5 & 79.9 & 75.9 & 67.1 & 30 \\
Virchow & 1.5M & Private & 82.8 & 4.5 & 77.4 & 71.8 & 69.1 & 36 \\
\rowcolor{gray!10}
OpenMidnight & 12K & TCGA & 84.8 & 5.3 & 79.3 & 43.6 & 69.1 & 37 \\
Phikon2 & 60K & TCGA, GTEx, CPTAC, others & 79.7 & 3.8 & 73.9 & 71.8 & 67.3 & 39 \\
\midrule
\rowcolor{gray!10}
\multicolumn{9}{l}{\textit{Natural-image models}} \\
\midrule
DINOv3-L & --- & --- & 77.9 & 3.0 & 71.5 & 61.8 & 61.6 & 44 \\
\rowcolor{gray!10}
DINOv2-L & --- & --- & 77.3 & 4.5 & 70.7 & 58.8 & 59.7 & 55 \\
DINOv2-B & --- & --- & 76.5 & 5.3 & 69.1 & 60.2 & 60.0 & 57 \\
\bottomrule
\end{tabular}%
}
\par\smallskip
\begin{minipage}{\linewidth}
\footnotesize
\textsuperscript{\dag}H0-mini was distilled on TCGA from H-Optimus-0, which was pretrained on private data.
\end{minipage}
\end{table}

\subsection{Ablation Studies} 
\label{sec:ablation}

We investigate four aspects of the training recipe: DINOv2 initialization, pretraining data volume, model size, and tissue composition. These experiments were conducted using the OpenMidnight implementation before correction of the KDE regularizer’s sign and are evaluated on EVA. Appendix~\ref{app:kde_comparison} reports an additional comparison on EVA and HEST between models trained with the uncorrected and corrected KDE regularizer. The combined average score changes from 0.7704 to 0.7689, an absolute difference of 0.0015, indicating a small aggregate change in downstream performance following the sign correction.

\subsubsection{Effect of Initialization with Pretrained DINOv2 Weights}
\label{sec:init}

To investigate the effect of backbone initialization, we compare OpenMidnight ViT-G/14 runs initialized from publicly available DINOv2 weights or random weights. Both runs share the same training configuration. We periodically evaluate checkpoints on EVA to track downstream performance throughout pretraining.

\begin{figure}[!htbp]
    \centering    \includegraphics[width=0.7\textwidth]{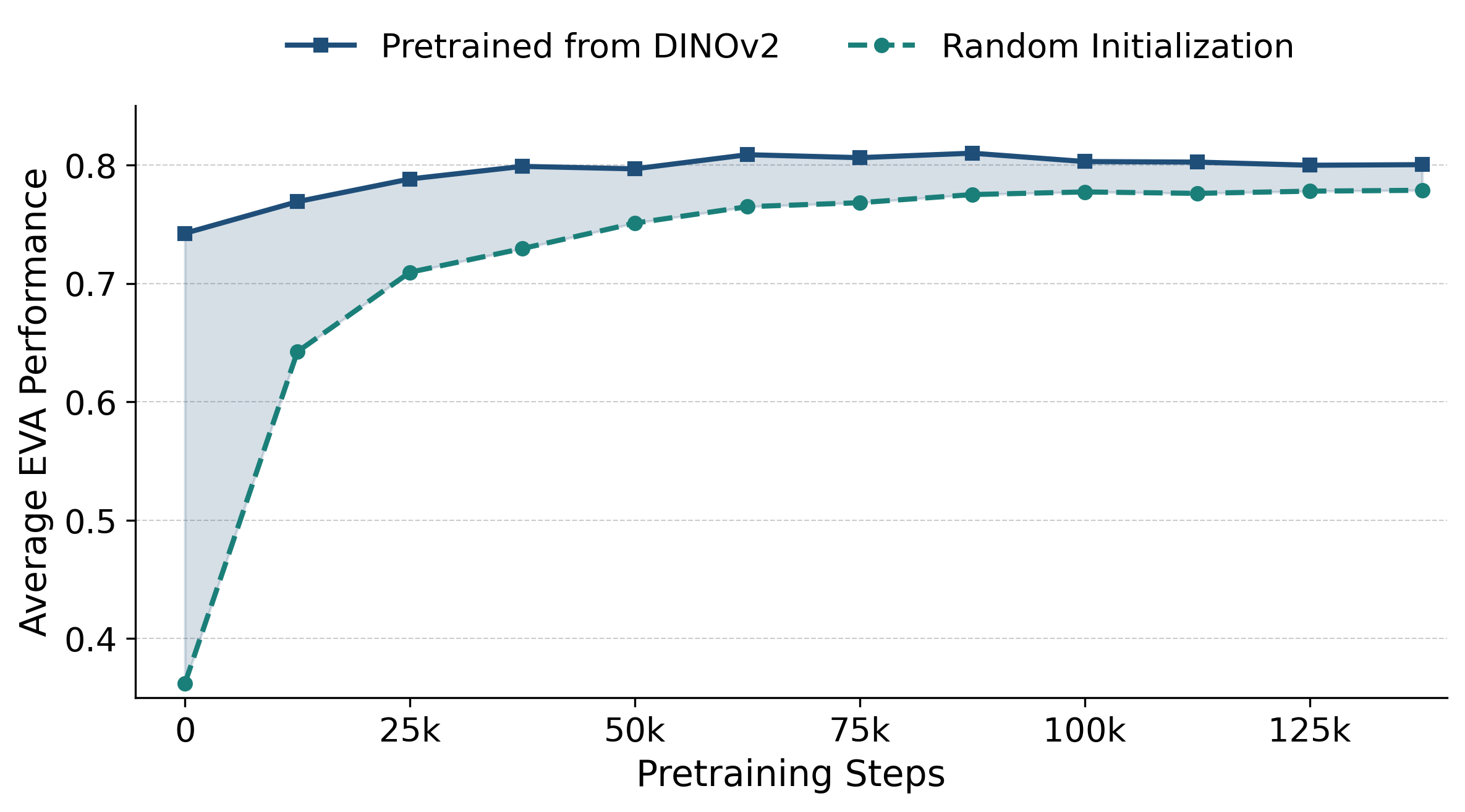}
    \caption{\textbf{Effect of pretrained initialization on downstream performance.} Mean EVA performance across pretraining checkpoints for OpenMidnight ViT-G/14 initialized from DINOv2 weights (solid) or random weights (dashed). Both runs use the same training configuration; shading indicates the performance gap.}
    \label{fig:scratch_vs_pretrained}
\end{figure}

DINOv2 initialization provides an early advantage: before pathology pretraining, mean EVA performance is 0.74 for the pretrained encoder and 0.37 for the randomly initialized encoder (Figure~\ref{fig:scratch_vs_pretrained}). The gap narrows during training; at 137,500 steps, the corresponding scores are 0.80 and 0.78. Thus, pretrained initialization improves performance throughout the observed training trajectory under the shared configuration. Using DINOv2 pretrained weights therefore offers a practical way to reduce the pathology-pretraining compute needed to achieve strong downstream performance, while separate hyperparameter tuning could further improve the randomly initialized model.

\subsubsection{Effect of Pretraining Data Volume}

To examine how pretraining data volume affects downstream performance, we compare OpenMidnight trained on the full TCGA corpus with a model trained on a randomly sampled 25\% subset of slides, holding the remaining training settings constant. Despite a fourfold reduction in pretraining slides, mean EVA performance decreases by only 0.008, from 0.815 to 0.807 (Table~\ref{tab:data_scaling}). This modest gap suggests limited gains from additional slides within the tested range, consistent with our broader observation that strong performance is possible without a massive pretraining corpus.
\begin{table}[htbp]
\centering
\small
\caption{EVA performance when pretraining on the full TCGA corpus versus a randomly sampled 25\% subset of slides. Classification tasks report balanced accuracy and segmentation tasks report Dice.}
\label{tab:data_scaling}
\resizebox{\textwidth}{!}{%
\begin{tabular}{lccccccccccc}
\toprule
\textbf{Data} & \textbf{BACH} & \textbf{BRACS} & \textbf{BreakHis} & \textbf{CRC} & \textbf{Gleason} & \textbf{MHIST} & \textbf{PCam-10} & \textbf{PCam} & \textbf{CoNSeP} & \textbf{MoNuSAC} & \textbf{Avg} \\
\midrule
100\% & 0.888 & 0.666 & 0.950 & 0.963 & 0.799 & 0.855 & 0.797 & 0.929 & 0.634 & 0.666 & 0.815 \\
25\%  & 0.900 & 0.652 & 0.879 & 0.963 & 0.789 & 0.840 & 0.831 & 0.929 & 0.638 & 0.653 & 0.807 \\
\bottomrule
\end{tabular}
}
\end{table}

\subsubsection{Effect of Model Sizes}
\label{sec:model_size_lr}

To examine whether increasing model size improves downstream performance, we compare OpenMidnight models with ViT-S, ViT-B, ViT-L, and ViT-G backbones, each initialized from the corresponding pretrained DINOv2 weights. Learning rate and stochastic depth rate were selected for each model size (Table~\ref{tab:model_configs}), while the remaining training hyperparameters were held constant.

\begin{figure}[htbp]
    \centering
    \includegraphics[width=0.49\linewidth]{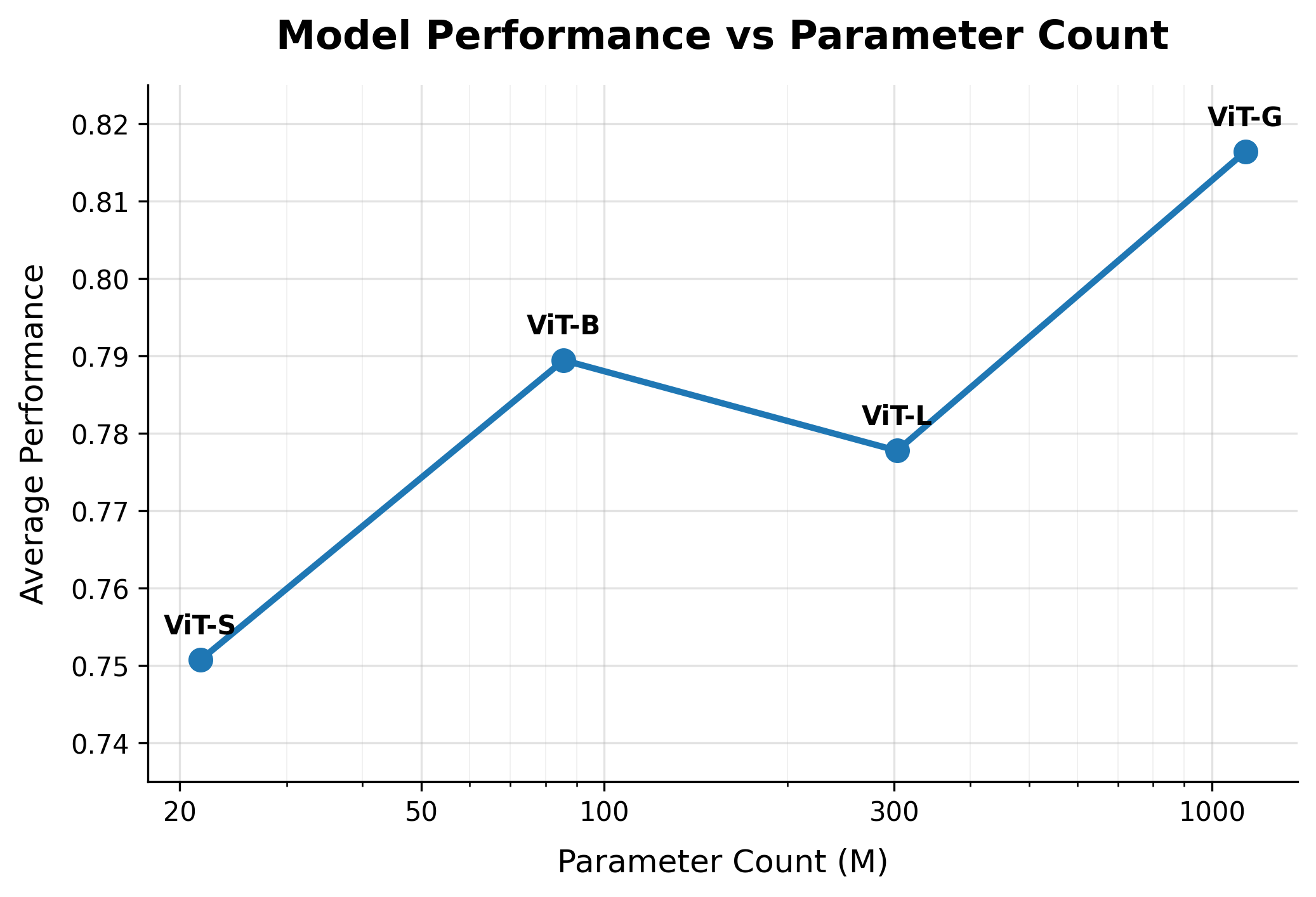}
    \hfill
    \includegraphics[width=0.49\linewidth]{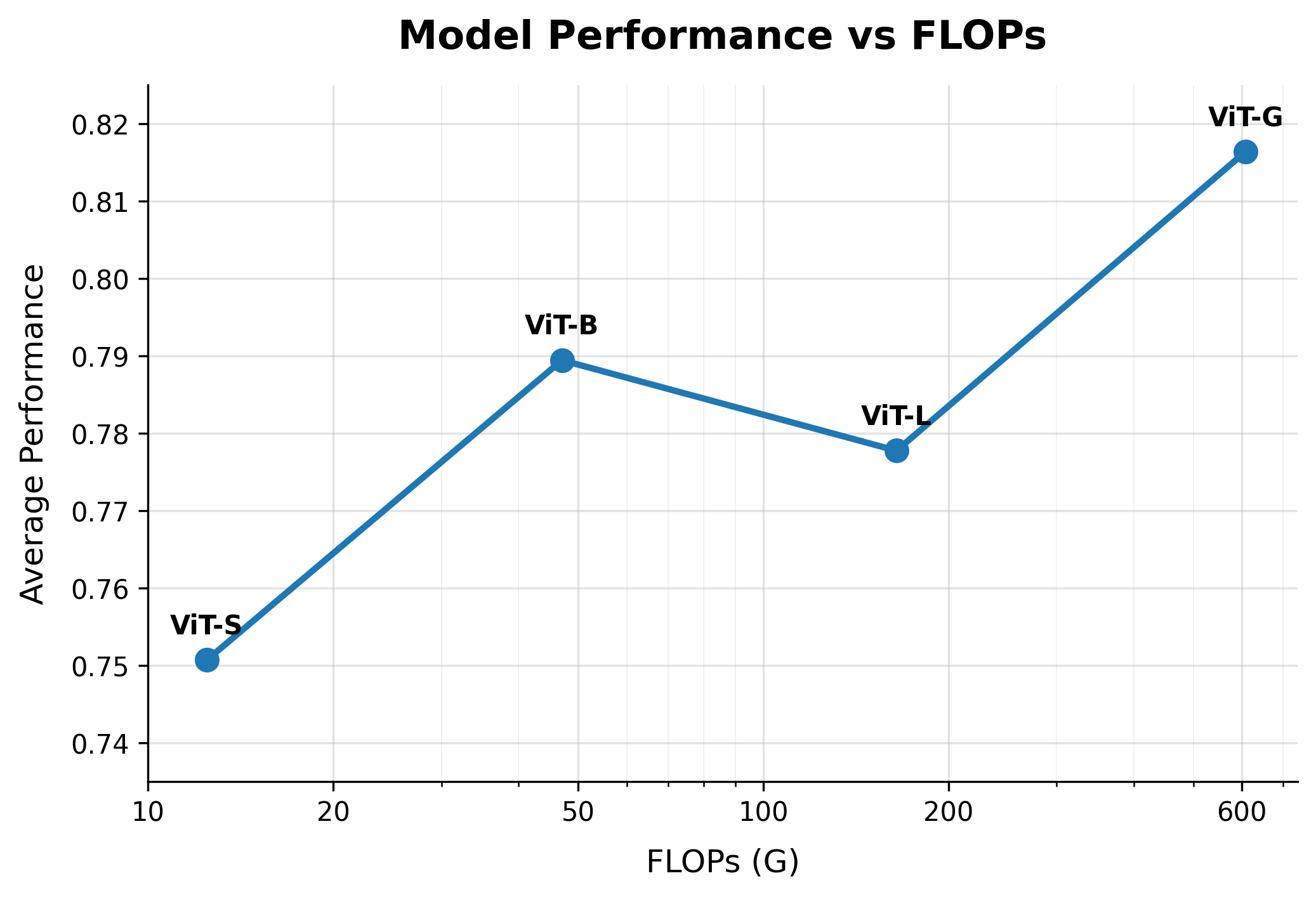}
    \caption{\textbf{Effect of model size on downstream performance.} Mean EVA benchmark performance for ViT-S, ViT-B, ViT-L, and ViT-G models, plotted against parameter count (left) and FLOPs (right), with logarithmic x-axes. All models are initialized with pretrained DINOv2 weights and use the model-specific learning rates and stochastic depth rates listed in Table~\ref{tab:model_configs}.}
    \label{fig:model_size_exp}
\end{figure}

Mean EVA performance generally increases with model size, although the trend is not monotonic: ViT-L scores 0.778, below ViT-B at 0.789 (Figure~\ref{fig:model_size_exp}). Further experiments with more extensive hyperparameter tuning for each model size are needed to clarify this relationship. ViT-G nevertheless achieves the highest mean EVA score (0.816), suggesting that greater model capacity may improve downstream performance under the tested training conditions.
\begin{table}[htbp]
\centering
\caption{Learning rate and drop path rate for each model size.}
\label{tab:model_configs}
\begin{tabular}{lcc}
\toprule
\textbf{Model} & \textbf{Learning rate} & \textbf{Drop path rate} \\
\midrule
ViT-G & $3 \times 10^{-4}$ & 0.4 \\
ViT-L & $4 \times 10^{-4}$ & 0.2 \\
ViT-B & $5 \times 10^{-4}$ & 0.2 \\
ViT-S & $7 \times 10^{-4}$ & 0.0 \\
\bottomrule
\end{tabular}
\end{table}

\subsubsection{Effect of Pretraining Tissue Composition}

To investigate whether pretraining tissue composition affects related downstream tasks, we exclude all breast slides from the TCGA corpus and compare the resulting OpenMidnight model with a full-corpus baseline under the same training setup. Breast slides account for approximately 10\% of the pretraining corpus (Figure~\ref{fig:data_distribution}). We group EVA tasks into breast-related tasks (BACH, BRACS, BreakHis, PCam, PCam-10) and other tasks (CRC, Gleason, MHIST, CoNSeP, MoNuSAC)\footnote{MoNuSAC includes samples from multiple tissue types; we group it with the other tasks for this analysis.}.

\begin{figure}[htbp]
    \centering
    {\small\textcolor[HTML]{22547D}{\rule{1.2em}{0.7em}}~OpenMidnight\qquad\textcolor[HTML]{1D817A}{\rule{1.2em}{0.7em}}~OpenMidnight (breast excluded)\par}
    \vspace{0.2em}
    \includegraphics[width=0.9\textwidth,trim=0 0 0 25,clip]{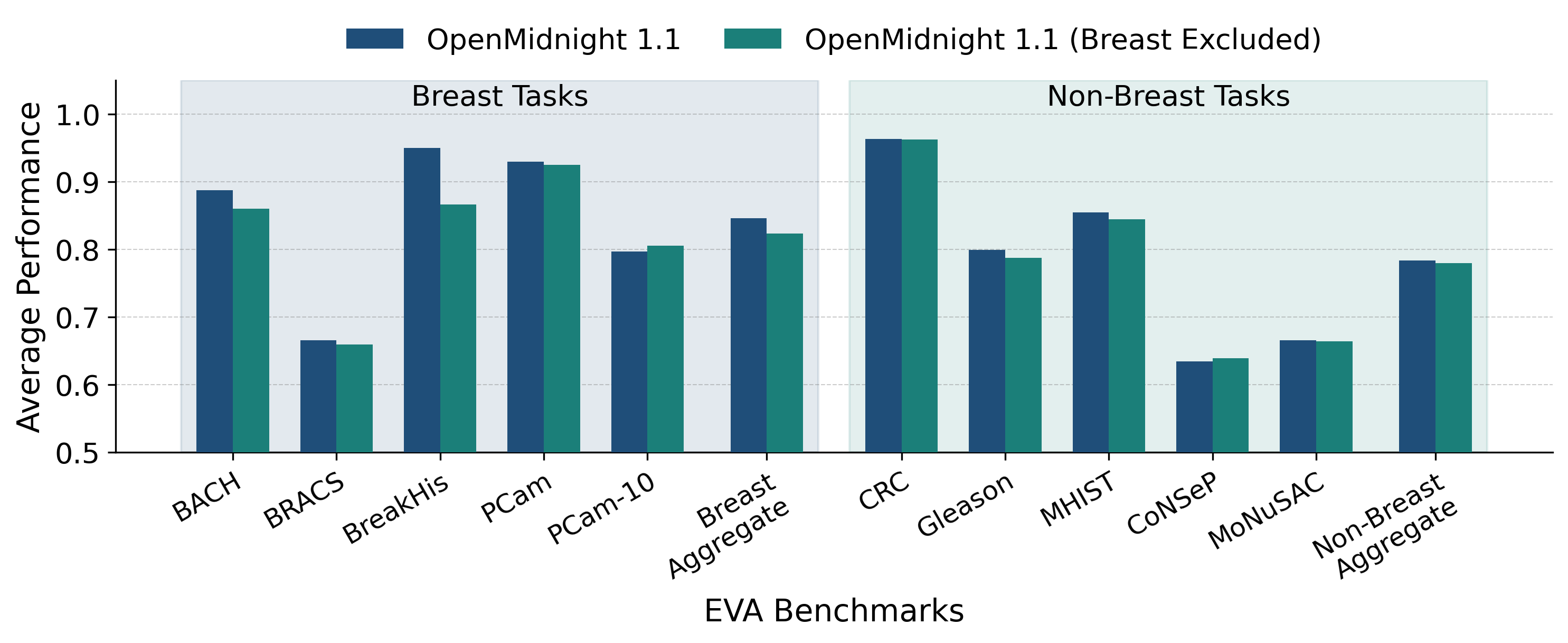}
    \caption{\textbf{Effect of excluding breast tissue from pretraining.} EVA performance of OpenMidnight and an otherwise matched model trained without breast WSIs. Tasks are grouped as breast-related (left) and other tasks (right).}
    \label{fig:breast_inclusion}
\end{figure}

Mean performance on breast-related tasks decreases from 0.845 to 0.824 when breast slides are excluded (Figure~\ref{fig:breast_inclusion}), while the mean for the other tasks changes from 0.784 to 0.780, with individual scores changing by approximately 0.01 or less. The larger decline in the breast-related group is consistent with a contribution from tissue-matched pretraining data. The stable non-breast results indicate that the breast decline may be attributed to the missing breast tissue rather than a general loss from training on less data.

The breast-related decline is concentrated in BreakHis (from approximately 0.95 to 0.87) and, to a lesser extent, BACH (from approximately 0.89 to 0.86). BRACS and PCam change little, while PCam-10 improves slightly. PCam contains lymph-node sections with metastatic breast cancer rather than breast parenchyma, which may partly explain its stability. BRACS, on the other hand, had the lowest baseline score among breast tasks (approximately 0.66). This may suggest that its baseline performance relies more on morphological features shared across tissues than on breast-specific features.

\subsection{Qualitative Analysis of Representations and Prediction Errors}

\begin{figure}[!htpb]
    \centering
    \includegraphics[width=\linewidth]{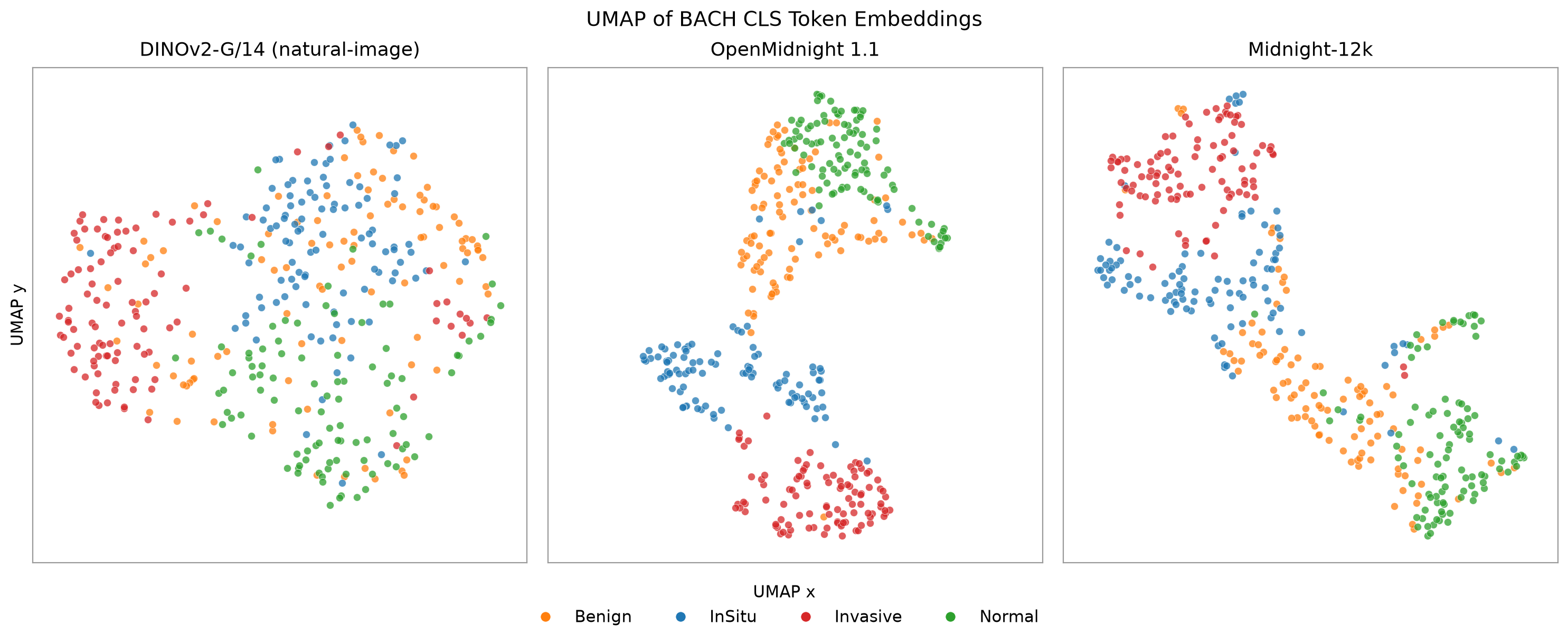}
    \caption{\textbf{Embedding-space geometry before and after pathology pretraining.} UMAP projections of CLS-token embeddings from all 400 BACH images for pretrained DINOv2-G/14, OpenMidnight 1.1, and Midnight. Points represent tiles and are colored by diagnostic category. Compared with DINOv2, the pathology-pretrained models show more distinct category-associated clustering.}
    \label{fig:bach_umap}
\end{figure}

To examine how pathology pretraining changes the geometry of the embedding space, we use BACH as a case study and compare representations from pretrained DINOv2-G/14, OpenMidnight 1.1, and Midnight. We project CLS-token embeddings from all 400 BACH images into two dimensions using UMAP, coloring each image by its category: normal, benign, in situ carcinoma, or invasive carcinoma (Figure~\ref{fig:bach_umap}). The DINOv2 embeddings show substantial overlap among categories, whereas the pathology-pretrained models exhibit more distinct category-associated clustering. In particular, OpenMidnight 1.1 and Midnight show clearer separation between in situ and invasive carcinoma, while normal and benign cases remain more intermixed. This visual pattern is consistent with their BACH balanced accuracies of 0.888, 0.904, and 0.725 for OpenMidnight 1.1, Midnight, and DINOv2-G/14, respectively.

\begin{figure}[!htpb]
    \centering
    \includegraphics[width=0.75\linewidth]{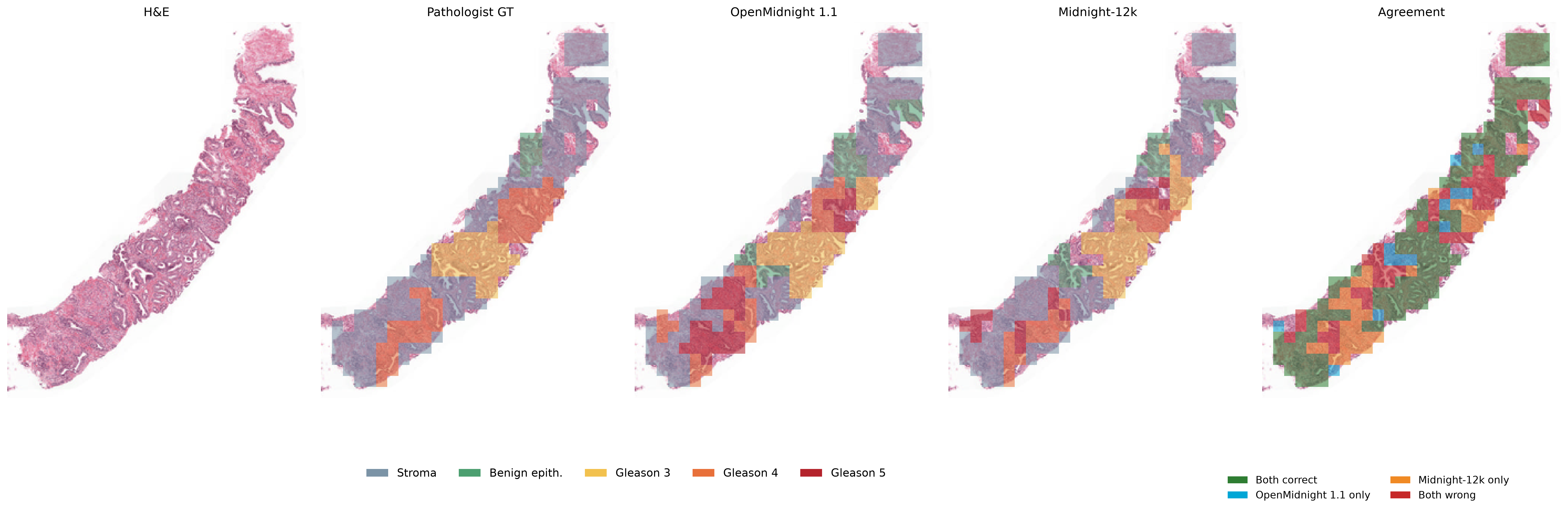}
    \vspace{-0.5em}
    \includegraphics[width=0.75\linewidth]{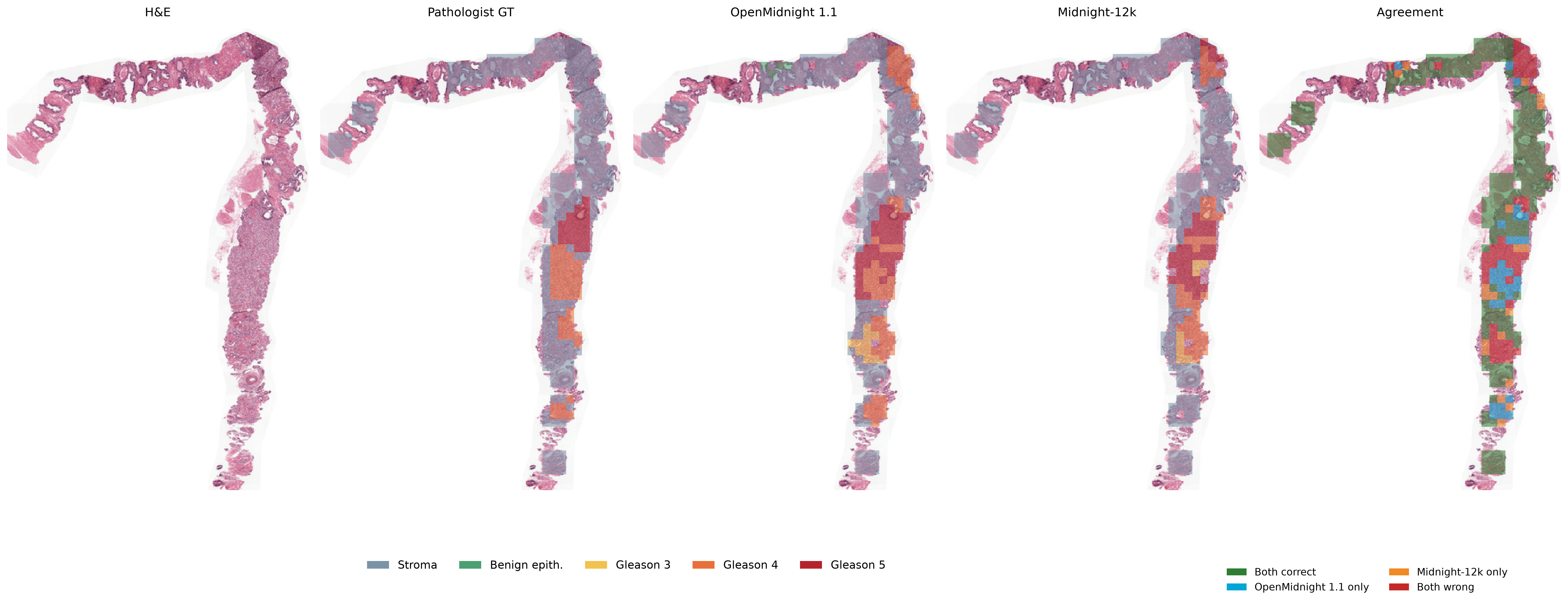}
    \caption{\textbf{Qualitative comparison of downstream segmentation errors on PANDA slides.} Tile-wise predictions from OpenMidnight 1.1 and Midnight on two selected PANDA slides. Each example shows the raw WSI, pathologist annotation, model predictions, and agreement with the annotation: both models correct, only OpenMidnight 1.1 correct, only Midnight correct, or both models incorrect. Predictions are based on local tile morphology and assembled into slide-level maps for visualization.}
    \label{fig:pandas}
\end{figure}

To qualitatively examine errors when using different foundation models for downstream segmentation, we compare tile-wise Gleason-pattern predictions from OpenMidnight 1.1 and Midnight on two selected PANDA slides~\citep{bulten2022panda} (Figure~\ref{fig:pandas}). Each example displays the slide, pathologist annotation, model predictions, and their agreement with the annotation. Both models recover broad tissue regions but show local disagreements with the annotations, including confusion between stroma and tumor and among Gleason patterns. The agreement maps reveal errors shared by the models and regions where only one model is correct. These qualitative analyses complement aggregate benchmark scores and provide additional insights to guide improvements in foundation model training and the use of these models for downstream segmentation tasks.

\section{Conclusion and Limitations}

OpenMidnight 1.1 achieves competitive downstream performance using approximately 12,000 publicly available TCGA slides and 320 H100 GPU-hours—approximately 34-fold fewer reported GPU-hours than Midnight. We release the training code, preprocessing pipeline, and model weights to support independent reproduction, inspection of training choices, and further development. Our ablations additionally examine how initialization, data volume, model size, and tissue composition affect downstream performance, providing empirical evidence to inform future training recipes.

Our study has several limitations. Hyperparameters were not independently optimized for every ablation condition, and comparisons based on individual pretraining runs do not establish the variability of the observed effects. The ablations also used the implementation before correction of the KDE regularizer's sign. Although our training procedure reduces compute requirements, repeated pretraining remains costly, limiting the practical scope of extensive ablations and hyperparameter searches. Smaller experimental frameworks such as nanopath~\citep{Scotti2026nanopath} offer a complementary approach for rapidly exploring training choices before validating promising configurations at larger scale. Nonetheless, OpenMidnight 1.1 lowers the computational and access barriers to developing competitive pathology foundation models, enabling the community to reproduce, investigate, and improve their training using public data.

\section{Broader Impact Statement}
 
Pathology foundation models may support more efficient tissue analysis and research on clinically relevant outcomes. OpenMidnight 1.1 lowers barriers to this research through public training data, reduced pretraining compute, and openly released code and weights, enabling independent reproduction and further development. However, downstream errors and uneven performance across patient populations, institutions, and imaging conditions could cause harm if these models are used without adequate validation. OpenMidnight 1.1 is a research model, not a validated diagnostic system. Clinical applications require independent validation in the intended setting, assessment of subgroup performance and robustness, and appropriate clinical oversight. Users must also respect patient privacy and applicable data-use conditions when adapting or evaluating the model.

\bibliography{main}
\bibliographystyle{tmlr}

\appendix
\section{Appendix}
\subsection{KDE Regularizer Sign Correction}
\label{app:kde_comparison}

Table~\ref{tab:kde_comparison} compares the original OpenMidnight run with OpenMidnight 1.1 under otherwise identical training conditions. The average EVA/HEST score changes from 0.7704 to 0.7689, a difference of 0.0015 in this comparison.

\begin{table}[H]
\centering
\caption{Comparison of OpenMidnight (uncorrected KDE) and OpenMidnight 1.1 (corrected KDE) on EVA and HEST. Training hyperparameters are otherwise identical; averages include whole-slide tasks.}
\label{tab:kde_comparison}
\resizebox{\textwidth}{!}{%
\begin{tabular}{lcccccccccccccc}
\toprule
& BACH & BRACS & BreakHis & CRC & Gleason & MHIST & PCam-10 & PCam & CoNSeP & MoNuSAC & CAM16 & PANDA & HEST & \textbf{Avg} \\
\midrule
OpenMidnight & 0.8913 & 0.6631 & 0.9527 & 0.9636 & 0.7988 & 0.8547 & 0.7967 & 0.9293 & 0.6341 & 0.6613 & 0.8291 & 0.6565 & 0.3844 & \textbf{0.7704} \\
OpenMidnight 1.1 & 0.8881 & 0.6281 & 0.9635 & 0.9671 & 0.7888 & 0.8286 & 0.8075 & 0.9257 & 0.6418 & 0.6695 & 0.8461 & 0.6577 & 0.3834 & \textbf{0.7689} \\
\bottomrule
\end{tabular}%
}
\end{table}

\clearpage\subsection{Complete THUNDER Benchmark Results}\suppressfloats[t]
\label{app:thunder_full}

\ifdefined\section
\else
  \documentclass[10pt]{article}
  \usepackage[letterpaper,margin=1in]{geometry}
  \usepackage{booktabs,graphicx,xcolor,colortbl}
  \newcommand{\OpenMidnightStandaloneLinearprobing}{}
  \begin{document}
  \pagestyle{empty}
\fi

\begin{table}[!htbp]
\centering
\caption{\textbf{Linear probing.} Per-dataset F1 scores (\%) and their mean over 16 classification datasets; higher is better. Bold marks the best score in each dataset and the best mean score; OpenMidnight 1.1 is highlighted.}
\label{tab:thunder_linear_probing}
\small
\setlength{\tabcolsep}{3pt}
\renewcommand{\arraystretch}{1.05}
\resizebox{\linewidth}{!}{%
\begin{tabular}{l|rrrrrrrrrrrrrrrr|r}
\toprule
 & \multicolumn{16}{c}{\textbf{Linear probing (F1)}} & \\
\cmidrule(lr){2-17}
\textbf{Model} & \textbf{bach} & \textbf{bracs} & \textbf{break-h} & \textbf{ccrcc} & \textbf{crc} & \textbf{esca} & \textbf{mhist} & \textbf{pcam} & \textbf{tc-crc} & \textbf{tc-tils} & \textbf{tc-unif} & \textbf{wilds} & \textbf{spi-br} & \textbf{spi-co} & \textbf{spi-sk} & \textbf{spi-th} & \textbf{Avg} \\
\midrule
\rowcolor{gray!10}
\multicolumn{18}{l}{\textit{Histopathology models}} \\
\midrule
UNI2-H & 84.7 & \textbf{65.4} & 78.7 & 89.5 & 94.1 & 86.4 & 78.3 & \textbf{95.0} & 66.9 & 90.4 & 78.8 & \textbf{98.9} & 86.7 & 90.5 & 92.5 & \textbf{95.1} & \textbf{85.7} \\
\rowcolor{gray!10}
GenBio-PathFM & 81.6 & 60.4 & 74.8 & 91.6 & \textbf{95.6} & 84.7 & 83.2 & 93.8 & 68.0 & 90.6 & 79.9 & 98.2 & 86.0 & 90.5 & \textbf{93.5} & 94.6 & 85.4 \\
\rowcolor{blue!15}
\textbf{OpenMidnight 1.1} & 85.0 & 61.1 & \textbf{87.4} & 88.7 & 94.8 & 80.1 & 81.5 & 92.6 & 63.4 & 90.6 & 80.3 & 98.2 & 86.1 & 89.1 & 90.1 & 94.2 & 85.2 \\
\rowcolor{gray!10}
OpenMidnight & 84.6 & 59.4 & 77.3 & 92.0 & 94.5 & 79.8 & 81.9 & 93.2 & 66.3 & 91.0 & 82.7 & 96.5 & 84.6 & 88.3 & 90.5 & 93.6 & 84.8 \\
Midnight & \textbf{87.9} & 63.9 & 56.7 & 90.8 & 95.5 & 86.2 & 80.3 & 93.5 & 65.6 & 91.0 & \textbf{85.3} & 98.3 & 86.1 & 89.6 & 91.0 & 94.4 & 84.7 \\
\rowcolor{gray!10}
Virchow2 & 76.5 & 62.2 & 72.5 & \textbf{93.6} & 92.5 & \textbf{88.9} & 83.4 & 92.8 & \textbf{69.6} & 90.9 & 78.4 & 89.8 & \textbf{87.2} & \textbf{90.8} & 92.0 & 93.9 & 84.7 \\
H-Optimus-0 & 69.7 & 61.1 & 66.0 & 90.8 & 93.5 & 83.5 & 83.3 & 93.0 & 67.3 & 91.0 & 79.3 & 98.7 & \textbf{87.2} & 89.9 & 91.9 & 94.4 & 83.8 \\
\rowcolor{gray!10}
H0-mini & 71.3 & 60.6 & 69.4 & 90.2 & 95.4 & 84.4 & 78.7 & 94.5 & 66.5 & \textbf{91.7} & 78.2 & 98.2 & 86.0 & 90.2 & 90.4 & 93.4 & 83.7 \\
GigaPath & 71.8 & 62.2 & 71.6 & 90.3 & 94.6 & 81.8 & 80.4 & 93.0 & 68.3 & 89.7 & 75.2 & 97.5 & 85.8 & 90.3 & 91.9 & 92.6 & 83.6 \\
\rowcolor{gray!10}
Virchow & 66.2 & 59.9 & 62.7 & 91.7 & 93.7 & 84.3 & 82.4 & 93.6 & 65.5 & 90.7 & 74.8 & 97.6 & 86.2 & 90.2 & 91.3 & 94.7 & 82.8 \\
Phikon2 & 64.7 & 58.3 & 53.5 & 76.7 & 92.0 & 77.3 & 79.1 & 90.8 & 61.1 & 91.0 & 77.7 & 95.8 & 86.0 & 89.7 & 87.2 & 94.6 & 79.7 \\
\midrule
\rowcolor{gray!10}
\multicolumn{18}{l}{\textit{Natural-image models}} \\
\midrule
DINOv3-L & 57.9 & 54.6 & 77.3 & 82.5 & 91.5 & 76.0 & \textbf{84.1} & 88.8 & 55.6 & 87.0 & 59.3 & 94.8 & 78.3 & 86.3 & 82.5 & 90.1 & 77.9 \\
\rowcolor{gray!10}
DINOv2-L & 68.9 & 50.6 & 79.5 & 81.4 & 90.5 & 74.9 & 82.2 & 87.2 & 59.3 & 85.9 & 56.2 & 90.3 & 74.0 & 85.3 & 82.2 & 87.8 & 77.3 \\
DINOv2-B & 64.4 & 51.4 & 77.8 & 79.7 & 90.4 & 73.8 & 82.7 & 86.3 & 58.6 & 85.9 & 54.4 & 90.7 & 76.0 & 83.8 & 80.1 & 87.6 & 76.5 \\
\bottomrule
\end{tabular}%
}
\end{table}
\ifdefined\OpenMidnightStandaloneLinearprobing
  \def\OpenMidnightAppendixEnd{\end{document}}
\else
  \let\OpenMidnightAppendixEnd\relax
\fi
\OpenMidnightAppendixEnd

\ifdefined\section
\else
  \documentclass[10pt]{article}
  \usepackage[letterpaper,margin=1in]{geometry}
  \usepackage{booktabs,graphicx,xcolor,colortbl}
  \newcommand{\OpenMidnightStandaloneCalibration}{}
  \begin{document}
  \pagestyle{empty}
\fi

\begin{table}[!tbp]
\centering
\caption{\textbf{Prediction calibration.} Per-dataset ECE scores (\%) and their mean over 16 classification datasets; lower is better. Bold marks the best score in each dataset and the best mean score; OpenMidnight 1.1 is highlighted.}
\label{tab:thunder_calibration}
\small
\setlength{\tabcolsep}{3pt}
\renewcommand{\arraystretch}{1.05}
\resizebox{\linewidth}{!}{%
\begin{tabular}{l|rrrrrrrrrrrrrrrr|r}
\toprule
 & \multicolumn{16}{c}{\textbf{Prediction calibration (ECE)}} & \\
\cmidrule(lr){2-17}
\textbf{Model} & \textbf{bach} & \textbf{bracs} & \textbf{break-h} & \textbf{ccrcc} & \textbf{crc} & \textbf{esca} & \textbf{mhist} & \textbf{pcam} & \textbf{tc-crc} & \textbf{tc-tils} & \textbf{tc-unif} & \textbf{wilds} & \textbf{spi-br} & \textbf{spi-co} & \textbf{spi-sk} & \textbf{spi-th} & \textbf{Avg} \\
\midrule
\rowcolor{gray!10}
\multicolumn{18}{l}{\textit{Histopathology models}} \\
\midrule
Midnight & 5.6 & 12.1 & \textbf{2.1} & 3.5 & 2.7 & 0.9 & 2.1 & 0.7 & 2.5 & 0.8 & 5.6 & \textbf{0.2} & 1.7 & 1.1 & 3.2 & 1.5 & \textbf{2.9} \\
\rowcolor{gray!10}
GigaPath & 7.1 & 7.0 & 9.1 & 5.2 & \textbf{2.0} & 1.4 & 3.9 & 2.7 & 3.1 & 0.3 & 3.7 & 1.0 & 3.9 & 2.9 & 0.5 & 2.3 & 3.5 \\
Phikon2 & 5.7 & 7.6 & 17.2 & 9.3 & 4.8 & 0.8 & 2.5 & \textbf{0.6} & \textbf{0.9} & 0.8 & 0.8 & 1.6 & 2.2 & 2.4 & 2.5 & 0.5 & 3.8 \\
\rowcolor{gray!10}
H0-mini & 6.8 & 10.9 & 14.1 & 3.3 & 2.9 & 0.9 & 7.4 & 1.0 & 4.4 & \textbf{0.1} & 0.9 & 0.8 & 3.8 & 0.7 & 0.5 & 1.8 & 3.8 \\
UNI2-H & 8.4 & \textbf{6.2} & 6.9 & 8.1 & 3.8 & 1.2 & 7.2 & 1.2 & 5.0 & 1.1 & 3.6 & 1.0 & 5.8 & \textbf{0.3} & 2.0 & \textbf{0.4} & 3.9 \\
\rowcolor{gray!10}
Virchow2 & 9.7 & 6.7 & 5.7 & 4.3 & 4.1 & \textbf{0.5} & 1.2 & 4.3 & 5.5 & 0.3 & 7.2 & 7.1 & 3.1 & \textbf{0.3} & 0.9 & 2.7 & 4.0 \\
H-Optimus-0 & 13.3 & 14.5 & 2.9 & 6.2 & 4.1 & 2.9 & 1.6 & 2.9 & 3.1 & 0.5 & 3.1 & 1.9 & 3.4 & 1.1 & \textbf{0.3} & 2.0 & 4.0 \\
\rowcolor{gray!10}
Virchow & 20.3 & 11.5 & 12.8 & 2.0 & 3.8 & \textbf{0.5} & 3.4 & 1.5 & 2.6 & 0.8 & 5.8 & 0.6 & 4.8 & 0.5 & 1.0 & 0.5 & 4.5 \\
GenBio-PathFM & 18.1 & 16.1 & 9.3 & 12.1 & 2.6 & 2.0 & 1.5 & 1.6 & 2.8 & 1.4 & \textbf{0.5} & 0.3 & 6.7 & 1.3 & \textbf{0.3} & 0.6 & 4.8 \\
\rowcolor{gray!10}
OpenMidnight & 16.9 & 13.6 & 14.8 & 12.8 & 3.4 & 0.9 & 3.0 & 1.7 & 1.7 & 0.3 & 1.3 & 1.3 & 6.8 & 2.4 & 2.8 & 1.0 & 5.3 \\
\rowcolor{blue!15}
\textbf{OpenMidnight 1.1} & 18.8 & 15.2 & 7.7 & 22.2 & 3.3 & 3.4 & 4.1 & 2.1 & 4.7 & 0.4 & 1.8 & \textbf{0.2} & 3.4 & 2.3 & 2.1 & 0.6 & 5.8 \\
\midrule
\rowcolor{gray!10}
\multicolumn{18}{l}{\textit{Natural-image models}} \\
\midrule
DINOv3-L & 6.0 & 8.7 & 4.4 & \textbf{1.5} & 3.1 & 0.8 & 1.2 & 0.9 & 3.3 & 0.8 & 6.3 & 1.4 & 1.6 & 0.5 & 6.4 & 1.0 & 3.0 \\
\rowcolor{gray!10}
DINOv2-L & \textbf{4.6} & 10.2 & 2.6 & 5.1 & 5.4 & 2.5 & 6.2 & 2.6 & 2.6 & 1.2 & 10.0 & 4.5 & \textbf{1.1} & 1.5 & 5.6 & 5.8 & 4.5 \\
DINOv2-B & 16.1 & 24.6 & 6.8 & 8.3 & 3.1 & 1.1 & \textbf{0.6} & 3.4 & 2.0 & 1.0 & 2.2 & 1.2 & 6.2 & 1.7 & 4.7 & 1.4 & 5.3 \\
\bottomrule
\end{tabular}%
}
\end{table}
\ifdefined\OpenMidnightStandaloneCalibration
  \def\OpenMidnightAppendixEnd{\end{document}}
\else
  \let\OpenMidnightAppendixEnd\relax
\fi
\OpenMidnightAppendixEnd

\ifdefined\section
\else
  \documentclass[10pt]{article}
  \usepackage[letterpaper,margin=1in]{geometry}
  \usepackage{booktabs,graphicx,xcolor,colortbl}
  \newcommand{\OpenMidnightStandaloneKnn}{}
  \begin{document}
  \pagestyle{empty}
\fi

\begin{table}[!tbp]
\centering
\caption{\textbf{KNN classification.} Per-dataset F1 scores (\%) and their mean over 16 classification datasets; higher is better. Bold marks the best score in each dataset and the best mean score; OpenMidnight 1.1 is highlighted.}
\label{tab:thunder_knn}
\small
\setlength{\tabcolsep}{3pt}
\renewcommand{\arraystretch}{1.05}
\resizebox{\linewidth}{!}{%
\begin{tabular}{l|rrrrrrrrrrrrrrrr|r}
\toprule
 & \multicolumn{16}{c}{\textbf{KNN (F1)}} & \\
\cmidrule(lr){2-17}
\textbf{Model} & \textbf{bach} & \textbf{bracs} & \textbf{break-h} & \textbf{ccrcc} & \textbf{crc} & \textbf{esca} & \textbf{mhist} & \textbf{pcam} & \textbf{tc-crc} & \textbf{tc-tils} & \textbf{tc-unif} & \textbf{wilds} & \textbf{spi-br} & \textbf{spi-co} & \textbf{spi-sk} & \textbf{spi-th} & \textbf{Avg} \\
\midrule
\rowcolor{gray!10}
\multicolumn{18}{l}{\textit{Histopathology models}} \\
\midrule
GenBio-PathFM & \textbf{85.8} & 57.0 & \textbf{84.2} & \textbf{92.8} & 94.8 & 83.6 & 69.7 & 90.2 & 64.4 & 89.2 & 76.8 & 95.5 & 81.4 & \textbf{88.0} & \textbf{92.2} & 93.0 & \textbf{83.7} \\
\rowcolor{gray!10}
UNI2-H & 85.3 & 56.1 & 79.4 & 90.9 & \textbf{95.0} & 83.2 & 70.9 & \textbf{93.4} & 64.1 & 89.1 & 75.1 & \textbf{98.0} & \textbf{82.6} & 87.1 & 90.5 & 92.5 & 83.3 \\
Virchow2 & 81.6 & 54.9 & 80.1 & 91.1 & 94.3 & \textbf{86.6} & 73.9 & 89.7 & \textbf{65.6} & 88.7 & 70.9 & 97.6 & 82.3 & \textbf{88.0} & 89.1 & 92.6 & 82.9 \\
\rowcolor{gray!10}
H-Optimus-0 & 65.6 & 52.2 & 81.9 & 90.5 & 93.1 & 82.1 & 73.2 & 89.3 & 62.4 & 89.1 & 73.2 & 97.9 & 81.8 & 87.8 & 89.3 & \textbf{93.8} & 81.4 \\
\rowcolor{blue!15}
\textbf{OpenMidnight 1.1} & 74.5 & 53.9 & 83.7 & 91.8 & 94.6 & 79.6 & 70.6 & 86.0 & 59.4 & 89.2 & 72.7 & 93.1 & 80.1 & 85.9 & 85.5 & 92.5 & 80.8 \\
\rowcolor{gray!10}
Midnight & 84.3 & 50.2 & 58.1 & 91.6 & 94.2 & 81.4 & 69.6 & 88.0 & 61.9 & 87.7 & \textbf{77.4} & 95.0 & 77.1 & 85.0 & 85.7 & 92.7 & 80.0 \\
GigaPath & 72.1 & 53.6 & 68.3 & 87.1 & 93.3 & 78.0 & 70.3 & 91.4 & 61.7 & 87.5 & 66.7 & 97.8 & 82.4 & 87.7 & 87.5 & 92.3 & 79.9 \\
\rowcolor{gray!10}
H0-mini & 71.8 & \textbf{57.2} & 62.8 & 86.3 & 94.4 & 81.1 & 72.5 & 88.8 & 62.7 & \textbf{90.2} & 64.0 & 95.7 & 80.9 & \textbf{88.0} & 86.2 & 92.8 & 79.7 \\
OpenMidnight & 82.7 & 50.1 & 78.1 & 86.1 & 94.7 & 74.1 & 66.5 & 84.0 & 60.4 & 87.0 & 75.6 & 92.5 & 78.3 & 83.5 & 83.9 & 91.3 & 79.3 \\
\rowcolor{gray!10}
Virchow & 55.2 & 51.7 & 56.0 & 83.8 & 92.5 & 81.8 & 68.2 & 90.0 & 61.8 & 88.7 & 62.7 & 97.5 & 79.3 & 87.8 & 88.8 & 92.3 & 77.4 \\
Phikon2 & 53.1 & 45.5 & 51.6 & 77.2 & 92.1 & 75.9 & 66.5 & 82.2 & 56.8 & 80.8 & 69.1 & 91.0 & 80.2 & 86.1 & 83.3 & 91.4 & 73.9 \\
\midrule
\rowcolor{gray!10}
\multicolumn{18}{l}{\textit{Natural-image models}} \\
\midrule
DINOv3-L & 60.9 & 43.8 & 71.9 & 76.6 & 88.3 & 70.1 & 73.1 & 81.6 & 58.9 & 83.8 & 41.6 & 91.8 & 66.9 & 79.8 & 73.1 & 81.6 & 71.5 \\
\rowcolor{gray!10}
DINOv2-L & 63.4 & 47.4 & 65.0 & 74.2 & 86.3 & 68.3 & \textbf{80.5} & 80.8 & 58.0 & 83.7 & 38.0 & 90.3 & 66.0 & 79.2 & 71.4 & 78.4 & 70.7 \\
DINOv2-B & 59.7 & 45.6 & 64.1 & 73.1 & 86.7 & 66.7 & 74.7 & 80.6 & 58.2 & 82.5 & 36.5 & 87.6 & 64.0 & 77.4 & 70.4 & 77.6 & 69.1 \\
\bottomrule
\end{tabular}%
}
\end{table}
\ifdefined\OpenMidnightStandaloneKnn
  \def\OpenMidnightAppendixEnd{\end{document}}
\else
  \let\OpenMidnightAppendixEnd\relax
\fi
\OpenMidnightAppendixEnd

\ifdefined\section
\else
  \documentclass[10pt]{article}
  \usepackage[letterpaper,margin=1in]{geometry}
  \usepackage{booktabs,graphicx,xcolor,colortbl}
  \newcommand{\OpenMidnightStandaloneFewshot}{}
  \begin{document}
  \pagestyle{empty}
\fi

\begin{table}[!tbp]
\centering
\caption{\textbf{16-shot classification.} Per-dataset F1 scores (\%) and their mean over 16 classification datasets; higher is better. Bold marks the best score in each dataset and the best mean score; OpenMidnight 1.1 is highlighted.}
\label{tab:thunder_few_shot}
\small
\setlength{\tabcolsep}{3pt}
\renewcommand{\arraystretch}{1.05}
\resizebox{\linewidth}{!}{%
\begin{tabular}{l|rrrrrrrrrrrrrrrr|r}
\toprule
 & \multicolumn{16}{c}{\textbf{16-shot (F1)}} & \\
\cmidrule(lr){2-17}
\textbf{Model} & \textbf{bach} & \textbf{bracs} & \textbf{break-h} & \textbf{ccrcc} & \textbf{crc} & \textbf{esca} & \textbf{mhist} & \textbf{pcam} & \textbf{tc-crc} & \textbf{tc-tils} & \textbf{tc-unif} & \textbf{wilds} & \textbf{spi-br} & \textbf{spi-co} & \textbf{spi-sk} & \textbf{spi-th} & \textbf{Avg} \\
\midrule
\rowcolor{gray!10}
\multicolumn{18}{l}{\textit{Histopathology models}} \\
\midrule
UNI2-H & \textbf{84.7} & \textbf{54.7} & 69.9 & \textbf{90.6} & \textbf{95.5} & \textbf{76.7} & 68.1 & \textbf{92.9} & 60.1 & 82.5 & 69.2 & 95.9 & 75.0 & 86.0 & 85.0 & 89.9 & \textbf{79.8} \\
\rowcolor{gray!10}
GenBio-PathFM & 79.9 & 51.8 & 68.1 & 85.9 & 95.3 & 75.6 & 69.6 & 88.3 & 63.0 & 88.0 & \textbf{71.6} & 94.0 & 75.8 & 86.1 & \textbf{88.2} & 89.7 & 79.4 \\
H-Optimus-0 & 55.2 & 43.1 & 70.6 & 83.9 & 91.2 & 74.2 & 65.5 & 85.9 & 61.4 & 88.5 & 66.9 & 93.6 & \textbf{76.4} & 86.9 & 83.5 & \textbf{91.5} & 76.2 \\
\rowcolor{gray!10}
GigaPath & 59.7 & 45.6 & \textbf{74.9} & 82.9 & 90.1 & 72.8 & 65.1 & 92.4 & \textbf{65.1} & 83.2 & 58.7 & 95.1 & 70.2 & \textbf{87.5} & 80.1 & 90.6 & 75.9 \\
H0-mini & 66.0 & 46.9 & 63.5 & 86.8 & 94.9 & 69.4 & 65.4 & 84.6 & 57.9 & \textbf{89.7} & 55.8 & 93.9 & 72.5 & 85.8 & 76.8 & 89.9 & 75.0 \\
\rowcolor{blue!15}
\textbf{OpenMidnight 1.1} & 76.0 & 50.4 & 64.7 & 63.6 & 94.7 & 75.9 & 58.8 & 85.3 & 57.7 & 85.4 & 63.5 & 93.6 & 67.6 & 82.0 & 76.0 & 89.4 & 74.0 \\
Virchow2 & 80.7 & 51.3 & 59.4 & 79.9 & 91.3 & 70.0 & 68.5 & 80.7 & 56.4 & 81.6 & 57.4 & 94.2 & 70.3 & 76.7 & 79.2 & 85.8 & 74.0 \\
\rowcolor{gray!10}
Virchow & 49.6 & 40.1 & 50.5 & 88.3 & 90.3 & 73.7 & 60.0 & 88.2 & 57.0 & 76.9 & 50.7 & \textbf{96.8} & 75.6 & 84.0 & 79.8 & 87.8 & 71.8 \\
Phikon2 & 57.3 & 42.8 & 60.9 & 90.1 & 89.6 & 67.0 & 55.7 & 83.5 & 56.2 & 85.6 & 60.2 & 92.2 & 65.9 & 82.3 & 75.4 & 83.6 & 71.8 \\
\rowcolor{gray!10}
Midnight & 82.5 & 49.6 & 38.5 & 77.5 & 94.7 & 75.0 & 62.4 & 82.8 & 55.1 & 76.2 & 63.4 & 89.9 & 61.4 & 75.6 & 73.0 & 86.4 & 71.5 \\
OpenMidnight & 34.2 & 24.1 & 45.4 & 22.6 & 53.6 & 34.6 & 48.5 & 56.0 & 51.7 & 60.8 & 45.3 & 47.3 & 31.3 & 49.0 & 49.4 & 44.4 & 43.6 \\
\midrule
\rowcolor{gray!10}
\multicolumn{18}{l}{\textit{Natural-image models}} \\
\midrule
DINOv3-L & 56.9 & 40.5 & 59.0 & 75.2 & 80.4 & 51.1 & 72.9 & 73.7 & 52.3 & 65.5 & 30.6 & 86.3 & 52.4 & 67.6 & 58.3 & 66.5 & 61.8 \\
\rowcolor{gray!10}
DINOv2-B & 56.6 & 38.5 & 63.8 & 74.6 & 80.2 & 51.1 & 71.8 & 70.7 & 51.2 & 64.7 & 26.0 & 83.3 & 51.2 & 66.0 & 54.6 & 59.6 & 60.2 \\
DINOv2-L & 55.1 & 41.2 & 57.6 & 69.0 & 77.3 & 51.0 & \textbf{73.0} & 65.2 & 53.3 & 53.7 & 26.7 & 87.8 & 50.4 & 66.3 & 55.2 & 57.7 & 58.8 \\
\bottomrule
\end{tabular}%
}
\end{table}
\ifdefined\OpenMidnightStandaloneFewshot
  \def\OpenMidnightAppendixEnd{\end{document}}
\else
  \let\OpenMidnightAppendixEnd\relax
\fi
\OpenMidnightAppendixEnd

\ifdefined\section
\else
  \documentclass[10pt]{article}
  \usepackage[letterpaper,margin=1in]{geometry}
  \usepackage{booktabs,graphicx,xcolor,colortbl}
  \newcommand{\OpenMidnightStandaloneSegmentation}{}
  \begin{document}
  \pagestyle{empty}
\fi

\begin{table}[!tbp]
\centering
\caption{\textbf{Semantic segmentation.} Per-dataset Dice scores (\%) and their mean over four segmentation datasets; higher is better. Bold marks the best score in each dataset and the best mean score; OpenMidnight 1.1 is highlighted.}
\label{tab:thunder_segmentation}
\small
\setlength{\tabcolsep}{3pt}
\renewcommand{\arraystretch}{1.05}
\resizebox{\linewidth}{!}{%
\begin{tabular}{l|rrrr|r}
\toprule
 & \multicolumn{4}{c}{\textbf{Semantic segmentation (Dice)}} & \\
\cmidrule(lr){2-5}
\textbf{Model} & \textbf{OCELOT} & \textbf{PanNuke} & \textbf{SegPath-Ep} & \textbf{SegPath-Ly} & \textbf{Avg} \\
\midrule
\rowcolor{gray!10}
\multicolumn{6}{l}{\textit{Histopathology models}} \\
\midrule
\rowcolor{blue!15}
\textbf{OpenMidnight 1.1} & \textbf{81.1} & \textbf{63.0} & 71.1 & \textbf{63.7} & \textbf{69.7} \\
\rowcolor{gray!10}
UNI2-H & 80.8 & 62.0 & 71.0 & 62.9 & 69.2 \\
OpenMidnight & 80.0 & 62.3 & \textbf{71.4} & 62.9 & 69.1 \\
\rowcolor{gray!10}
Virchow & 80.3 & 62.6 & 70.6 & 62.8 & 69.1 \\
Virchow2 & 79.6 & 62.9 & 70.6 & 62.7 & 68.9 \\
\rowcolor{gray!10}
H0-mini & 76.8 & 62.0 & 71.2 & 62.7 & 68.2 \\
GenBio-PathFM & 79.7 & 61.6 & 69.8 & 61.5 & 68.2 \\
\rowcolor{gray!10}
Midnight & 77.0 & 61.8 & 70.5 & 62.7 & 68.0 \\
Phikon2 & 79.3 & 60.4 & 69.1 & 60.4 & 67.3 \\
\rowcolor{gray!10}
GigaPath & 77.9 & 60.4 & 68.5 & 61.7 & 67.1 \\
H-Optimus-0 & 75.4 & 55.4 & 69.3 & 58.2 & 64.6 \\
\midrule
\rowcolor{gray!10}
\multicolumn{6}{l}{\textit{Natural-image models}} \\
\midrule
DINOv3-L & 70.1 & 55.7 & 65.0 & 55.5 & 61.6 \\
\rowcolor{gray!10}
DINOv2-B & 72.2 & 46.7 & 63.7 & 57.2 & 60.0 \\
DINOv2-L & 71.6 & 46.3 & 64.0 & 57.0 & 59.7 \\
\bottomrule
\end{tabular}%
}
\end{table}
\ifdefined\OpenMidnightStandaloneSegmentation
  \def\OpenMidnightAppendixEnd{\end{document}}
\else
  \let\OpenMidnightAppendixEnd\relax
\fi
\OpenMidnightAppendixEnd

\end{document}